\documentclass{article}
\usepackage{spconf,amsmath,amssymb,graphicx,url}

\def\R{{\cal R}}

\title{WHAT IS THE COST OF ADDING A CONSTRAINT IN LINEAR LEAST SQUARES?}
\name{Ramakrishna Kakarala and Jun Wei\thanks{ram.kakarala,jun.wei@ovt.com}}
\address{Omnivision Technologies, Inc.}
\begin{document}
%
\maketitle
\begin{abstract}
    Although the theory of constrained least squares (CLS) estimation is well known, it is usually applied with the view that the constraints to be imposed are unavoidable.  However, there are cases in which constraints are optional.  For example, in camera color calibration, one of several possible color processing systems is obtained if a constraint on the row sums of a desired color correction matrix is imposed; in this example, it is not clear a priori whether imposing the constraint leads to better system performance.  In this paper, we derive an exact expression connecting the constraint to the increase in fitting error obtained from imposing it.  As another contribution, we show how to determine projection matrices that separate the measured data into two components: the first component drives up the fitting error due to imposing a constraint, and the second component is unaffected by the constraint.  We demonstrate the use of these results in the color calibration problem.
\end{abstract}
\section{Introduction}
\label{sec:intro}

This paper addresses the role of linear constraints in determining fitting error for linear least squares (LS) problems.  Specifically, we aim to theoretically determine the increase in fitting error due to imposing a linear constraint, and to determine the geometric relationship between constrained and unconstrained estimators.  Our motivation arises from a practical problem in color calibration, as we now describe.  

The color correction stage of an image signal processing system is used to convert an image sensor’s
color response into device-independent color values \cite{sharma03}.  This stage is typically performed
after Bayer color interpolation by applying a $3\times 3$ matrix,
known as the color correction matrix (CCM), to input red, green, and blue (RGB) values.  The corrected
values, written with subscript $c$, are then related to the input values as follows:

\begin{equation}
    \begin{bmatrix}
        R_c \\ G_c \\ B_c
    \end{bmatrix} =
    \begin{bmatrix}
        c_{11} & c_{12} & c_{13} \\
        c_{21} & c_{22} & c_{23} \\
        c_{31} & c_{32} & c_{33}
    \end{bmatrix}
    \begin{bmatrix}
        R \\
        G \\
        B
    \end{bmatrix}
\label{eq:ccmapply}
\end{equation}

\begin{figure}[htb]
    \centering
    \centerline{\includegraphics[width=6.5cm]{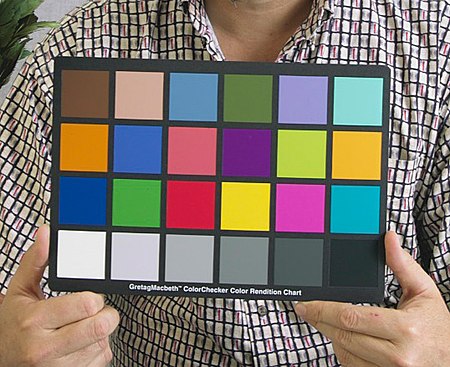}}
    \caption{Color checker chart, often referred to as a ``MacBeth'' chart, that is widely used for color calibration \cite{macbeth}.}
    \label{fig:macbeth}
\end{figure}

The problem of calibrating a CCM involves fitting a set of measured RGB vectors $m_i$, $i=1,\ldots,k$, to
a corresponding set of reference values $r_i$. For example, the $k=24$ color checker
chart shown in Figure~\ref{fig:macbeth} is used to measure RGB values from the sensor. These measurements are then compared to published reference values.  
Letting $M$ denote the $3\times k$ matrix whose columns are measured RGB values, and $R$ the corresponding reference matrix,  calibration requires minimizing 
over $3\times 3$ matrices $C$ the fitting error $\|CM-R\|^2$, where the norm is Frobenius.  This is a standard LS problem.  

However, in some systems, a CCM is applied after white balance. In this case, to preserve input neutral gray tones for which $R=G=B$, CCM matrices must be normalized so that their rows sum to $1.0$ \cite{bianco2013}.  Letting ${\bf 1}=[1,1,1]^\top$, we write a constrained linear least squares (CLS) problem as follows:
\begin{eqnarray}
    &{\rm minimize}\,\,\, \|CM - R\|^2&,\nonumber\\
    &{\rm subject\, to}\,\, C{\bf 1} = {\bf 1}&.
\label{eq:ccmfitting}
\end{eqnarray}
This row-sum constraint is not always used, as alternative color processing pipelines are possible without it \cite{sharma03}. 

It is clear that imposing any constraint on $C$ should increase fitting error, for constraints reduce the freedom in choosing $C$.   It is useful to have a mathematical model of fitting error due to linear constraints, which we provide in this paper.   Note that our goal is to analyze the CLS problem and contribute to its theory; the problem of CCM calibration is used only for illustrative purposes.  In particular, there exist CCM calibration algorithms besides least squares \cite{fang2017}, which we do not consider here.   

The general theory of solving CLS problems is well established \cite[Ch. 16]{boydvanden18}\cite[Ch. 1]{amemiya85}.  We review this theory, and point out simplifications that explicitly  separate the role of the unconstrained estimator from the constraint fitting error.  We determine the subspace on which both the constrained and the unconstrained estimators agree, and provide an exact result for the error increase due to applying the constraint.   We show how the simplified results apply to the CCM estimation problem, comparing our approach to others \cite{finlaysondrew97}\cite{wolftechreport}.  

\section{LS AND CLS ESTIMATION}

This section formulates LS and CLS problems as generally as possible, for real-valued vectors and matrices.  Let $M$ be a full-rank measurement matrix of arbitrary size (not necessarily $3\times k$ as in \S \ref{sec:intro}), $R$ a corresponding reference, and $W$ a positive definite weighting matrix for errors.  Define the objective function as 
\begin{equation}
\|CM-R\|^2_W = {\rm trace}\left\{\left(CM-R\right)W\left(CM-R\right)^{\top}\right\}.
\label{eq:objective}
\end{equation}
The unique minimizing solution is:
\begin{equation}
C_u = RWM^{\top} \left(M W M^{\top} \right)^{-1}
\label{eq:uLS}
\end{equation}
Henceforth, we use the subscript $u$ to indicate the solution to an unconstrained least-squares problem.

We now introduce linear constraints, formulating a general case that allows a variety of uses. Let $K$ denote a full row-rank matrix whose purpose is to impose a constraint on $C$ from the left, and similarly let $F$ be a full column-rank constraint matrix applied on the right, and finally let $G$ a matrix of requirements.  We write the general set of linear constraints as $KCF=G$.   Note that neither $K$ nor $F$ need to be square; for example, in the case where $C$ is $3\times 3$, we might have $K=[1,1,1]={\bf 1}^{\top}$, and $F={\bf 1}$.  The general CLS problem is now stated as follows: 
\begin{eqnarray}
    &{\rm minimize}\,\,\, \|CM - R\|^2_W,&\nonumber\\
    &{\rm subject\, to}\,\, KCF = G.&
\label{eq:CLS}
\end{eqnarray}
Henceforth, we denote the solution to this problem by $\widehat{C}$. 

We would expect that, if it so happened that the unconstrained estimate $C_u$ already met the constraint, i.e., $KC_uF = G$, then $\widehat{C}=C_u$ (the constrained estimate is the same as the unconstrained estimate).  Let $\Delta_c=KC_uF - G$ denote the error in meeting the constraint.  Intuitively, we would expect that, the smaller $\Delta_c$ is, the closer the constrained estimate $\widehat{C}$ is to $C_u$.  

Let us now derive a simple expression for $\widehat{C}$.  The standard theory \cite[Ch. 16]{boydvanden18}\cite[Ch. 1]{amemiya85} shows how to proceed when $C$ is a vector of parameters, rather than a matrix as we have formulated in (\ref{eq:CLS}).  We could, if we wish, reuse this theory by ``vectorizing'' (\ref{eq:CLS})  using the ${\rm vec}$ operator and the Kronecker product $\otimes$. Vectorization means the constraint in (\ref{eq:CLS}) is 
\begin{equation}
{\rm vec}\left(F^{\top}\otimes K\right) {\rm vec}(C) = {\rm vec} (G).
\label{eq:vecG}
\end{equation}
However, vectorization loses the intuition that comes from directly formulating the problem (\ref{eq:CLS}) using $C$ in matrix format, in the way $C$ is meant to be applied, as in eqn.  (\ref{eq:ccmapply}).  To preserve intuition, and to introduce useful simplifications, we follow \cite{boydvanden18}\cite{amemiya85} but work directly in matrix formulations. 

Letting $\Lambda$ be a matrix of Lagrange multipliers of the same size as $G$, we write the augmented function corresponding to (\ref{eq:CLS}) as follows:
\begin{equation}
\|CM - R\|^2_W + {\rm trace} \left\{\Lambda^{\top}\left(KCF-G\right)\right\} 
\end{equation}
Differentiating with respect to $C$ yields (using the rule \linebreak $\partial{\rm trace\left(AXB\right)}/\partial X = A^{\top} B^{\top}$) 
\begin{equation}
    2\left(\widehat{C}M- R\right)WM^{\top}+ K^{\top} \Lambda F^{\top} = 0.
    \label{eq:normal}
\end{equation}
Solving for $C$ yields
\begin{equation}
    \widehat{C}= \left(RWM^\top  - \frac{1}{2}K^{\top}\Lambda F^{\top}  \right) \left(M W M^{\top}\right)^{-1}
    \label{eq:solnlambda}
\end{equation}
Let $J=MWM^{\top}$; this matrix corresponds to the Fisher information matrix used in parameter estimation theory for the least squares problem \cite{gormanhero}. 
It is invertible due to our assumption that $M$ is full-rank and $W$ is positive definite.  
Setting $K\widehat{C}F=G$ and solving yields 
\begin{equation}
\Lambda = 2\left(KK^{\top}\right)^{-1} \Delta_c 
\left[F^{\top} J^{-1} F \right]^{-1}.
\label{eq:lambda}
\end{equation}
To write the solution, we construct for simplicity a left inverse of $F$, denoted $F_{\ell}$, by the matrix
\begin{equation}
F_{\ell} = \left(F^{\top} J^{-1} F \right)^{-1} F^{\top} J^{-1}.
\label{eq:Fleft}
\end{equation}
Note that $F_{\ell}F = I$.  Similarly, let $K_{\R}$ denote a right inverse of $K$; since $K$ is assumed to have full row rank, we use $K_{\R}=K^{\top} \left(KK^{\top}\right)^{-1}$.  Then we have our desired solution to the CLS problem:
\begin{equation}
\widehat{C} = C_u - K_{\R} \Delta_c F_{\ell}.
\label{eq:ourcls}
\end{equation}
We see the optimal constrained estimator $\widehat{C}$ corrects the unconstrained estimator $C_u$ by ``modulating'' the constraint error $\Delta_c=KC_uF - G$ with the inverses of the constraint matrices. It follows directly that this estimator meets the constraint:
\begin{equation}
K\widehat{C}F=KC_u F - KK_{\R} \Delta_c F_{\ell}F = G.
\label{eq:constraintmet}
\end{equation}

\subsection{A comparison}

We compare this derivation with an alternative approach to ``white-point preserving'' CCM estimation that is proposed by Finlayson and Drew \cite{finlaysondrew97}.  They consider the case where $W=I$, $K=I$, and $F$, $G$ are vectors denoted $f$, $g$, respectively.  They turn the CLS problem (\ref{eq:CLS})  into an unconstrained problem by writing $C=D+E$, where $D$ is any matrix such $Df=g$ (formed, for example by placing the ratios of $g$ to $f$ on the diagonal), and $E$  is a matrix such that $E f = 0$.  Letting $V$ denote a (rectangular) matrix whose columns form a basis for the hyperplane perpendicular to $f$, we set $E=NV^\top$ for a suitable rectangular matrix $N$. (Note that if $f$ is $n\times 1$, then both $V$ and $E$ are $n\times (n-1)$ matrices.) Now, we minimize over the {\em unconstrained} entries of $N$ the fitting error:
\begin{equation}
    \|(D+NV^T)M-R\|^2.
\end{equation}
The solution is shown \cite{finlaysondrew97} to be: 
\begin{equation}
    \widehat{C} = D + \left(R-DM\right)M^\top V \left(V^\top M M^\top V\right)^{-1}V^\top.
    \label{eq:finlaydrewsoln}
\end{equation}
As we discuss below, (\ref{eq:finlaydrewsoln}) produces exactly the same estimate as our solution (\ref{eq:ourcls}).  

\section{Main results}

We enumerate our main novel results in this section. 
\label{sec:obs} 
\begin{enumerate}
\item The estimator (\ref{eq:ourcls}) is the unique minimum of the CLS problem (\ref{eq:CLS}).  The Appendix contains a constructive proof; alternatively, we may prove using convex optimization theory \cite[Ch. 10]{bvconvex}.

\item  We show in the Appendix that the excess fitting error due to the constraint is:
          \begin{equation}
              \|\widehat{C}M - R\|^2_W = \| C_u M - R\|^2_W + \| K_{\R} \Delta_c F_{\ell}M \|^2_W.
              \label{eq:excesserror}
          \end{equation}
Note that the excess (second term on right) is independent of reference $R$.  

\item What component of the measurement $M$ contributes to the excess fitting error $\| K_{\R} \Delta_c F_{\ell}M \|^2_W$ in (\ref{eq:excesserror})? To answer this,  let us introduce two orthogonal projections $P_{\perp} = FF_{\ell}$ and $P_{||} = I-P_{\perp}$.  Then $P_{\perp}P_{\perp} = P_{\perp}$ and $P_{||}P_{||} = P_{||}$; also, $P_{||}P_{\perp} = 0$.  Therefore, 
\begin{equation}
M = P_{||}M  + P_{\perp}M.
\label{eq:projection}
\end{equation}
Since $F_{\ell} P_{\perp} = F_{\ell}$, and $F_{\ell}P_{||}=0$, we write the excess as 
\begin{equation}
K_{\R} \Delta_c F_{\ell}M = K_{\R} \Delta_c F_{\ell}P_{\perp}M.
\label{eq:excessparallel}
\end{equation}
Hence, only $P_{\perp}M$ contributes to increasing fitting error.   Note that both $P_{||}$ and $P_{\perp}$ depend only on the constraint matrix $F$, and are independent of $K$.

 \item From $F_{\ell}P_{||}=0$ and (\ref{eq:ourcls}), we see that, on $P_{||}M$, the constrained estimator $\widehat{C}$ and the unconstrained estimator $C_u$ must agree
\begin{equation}
 \widehat{C} P_{||}M = C_u P_{||} M.
\label{eq:agreeonperp}
\end{equation}

\item \label{item:crlb}
 Let us write the fitting problem as a constrained parameter vector estimation problem, using boldface to indicate vectorized matrix of the same letter, i.e., ${\bf x}={\rm vec }(X)$.  With $\eta$ representing Gaussian noise, $B=M^{\top} \otimes I$, and $A=F^{\top} \otimes K$, we write 
\begin{eqnarray}
    &{\bf r} = B {\bf c} + {\bf \eta}, &\nonumber\\
    &{\rm subject\, to}\,\,  A {\bf c} = {\bf g}.&
\label{eq:constraintfitting}
\end{eqnarray}
Gorman and Hero \cite{gormanhero} show that imposing constraints on ${\bf c}$ {\em reduces} the Cramer-Rao lower bound (CRLB) on estimator variance over the  unconstrained case.  Specifically, with $J_{u}=B^{\top} {\Sigma}^{-1} B$ representing the Fisher matrix of the unconstrained estimator,   the bound when constraints are imposed is:
\begin{equation}
{\rm Var}\{\widehat{\bf c}\} \geq J_{c}^{-1} := J_{u}^{-1} - J_u^{-1} A^{\top} \left[A J_u^{-1} A^{\top}\right]^{-1} A J_{u}^{-1}.
\end{equation}
Interestingly, even as constraints increase fitting error, they reduce CRLB. Furthermore, as Gross\cite[pg 94]{Gross03} shows, constraints also reduce mean squared error.     We may say that imposing constraints increases the precision of specifying the optimal estimator, but reduces the accuracy of that estimator.

\item The solution (\ref{eq:ourcls}) produces exactly the same matrix as (\ref{eq:finlaydrewsoln}), regardless of the choice of $D$ and $V$. This follows from the uniqueness of (\ref{eq:ourcls}) as the optimum solution of (\ref{eq:CLS}).  We may also prove the equality constructively by letting $D=ff_{\ell}$ in (\ref{eq:finlaydrewsoln}), and noting that we may write $(I-ff_{\ell})=NV^{\top}$ for a suitable rectangular matrix $N$.
   
\end{enumerate}

\section{EXAMPLE IN COLOR CALIBRATION}
\label{sec:exp}

For illustrative purposes, we apply the results of the previous section to color calibration.  We consider two optional constraints on CCM matrices. The first is the row-sum constraint described in \S~\ref{sec:intro}, that $C{\bf 1}={\bf 1}$.  This is obtained by setting $K=1.0$, $F={\bf 1}=G$.  The second requires that the sum of all entries of $C$ is $3.0$.  We write this as ${\bf 1}^{\top} C {\bf 1} = 3.0$, which means $K={\bf 1}^{\top}$, $F={\bf 1}$, and $G=3.0$.    Note that the row-sum constraint implies the total-sum constraint.

Let us compare the cost of the two constraints.  Since row-sum is stricter, we would expect its fitting error to be higher than for total-sum; correspondingly, we would also expect the CRLB to be lower for row-sum than for total-sum.  

For experimental purposes, we use an image of a measured chart 
\footnote{\url{https://www.imatest.com/wp-content/uploads/2011/11/Colorcheck_1_raw_1004W.jpg}}.  The image is white balanced to the white patch of the chart.   The coordinates of each of the $24$ patches in the chart are extracted using an automated tool \footnote{https://github.com/colour-science/colour-checker-detection}.  The mean patch  RGB values after white balance are used to construct a $3\times 24$ matrix $M$.  Correspondingly, we set $R$ from sRGB values \cite{pascale}.  In Table~\ref{tab:fiterr}, we compare fitting errors and CRLB matrices for both constraints. Results confirm the expectations described above, that fitting error increases with stricter constraints (in this example, by up to $17\%$, a significant amount), while CRLB decreases. 

\begin{figure}[htb]

\begin{minipage}[b]{0.48\linewidth}
  \centering
  \centerline{\includegraphics[width=4cm]{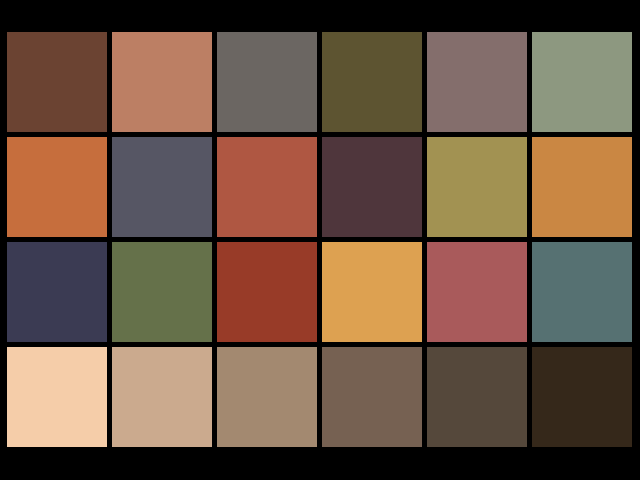}}
  \centerline{(a) Raw chart }\medskip
\end{minipage}
\hfill
\begin{minipage}[b]{0.48\linewidth}
\centering
  \centerline{\includegraphics[width=4cm]{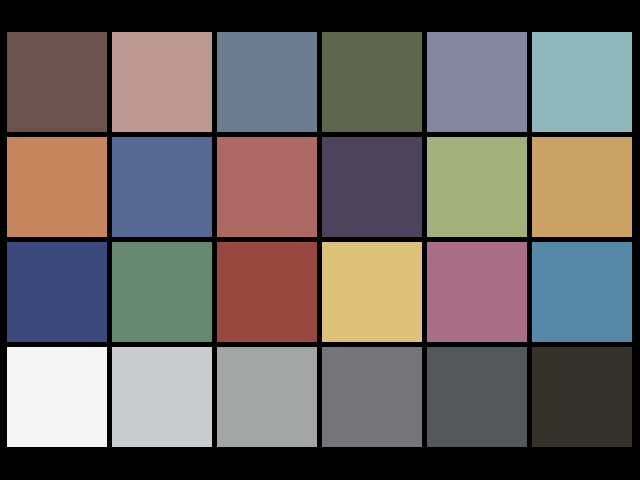}}
  \centerline{(b) After white balance (=$M$)}\medskip
\end{minipage}
\begin{minipage}[b]{0.48\linewidth}
  \centering
  \centerline{\includegraphics[width=4.0cm]{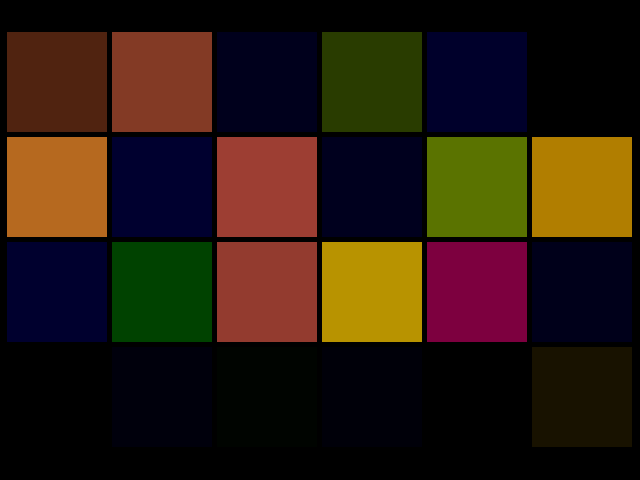}}
  \centerline{(c) Agreement ($P_{||}M$) chart }\medskip
\end{minipage}
\hfill
\begin{minipage}[b]{.48\linewidth}
  \centering
  \centerline{\includegraphics[width=4.0cm]{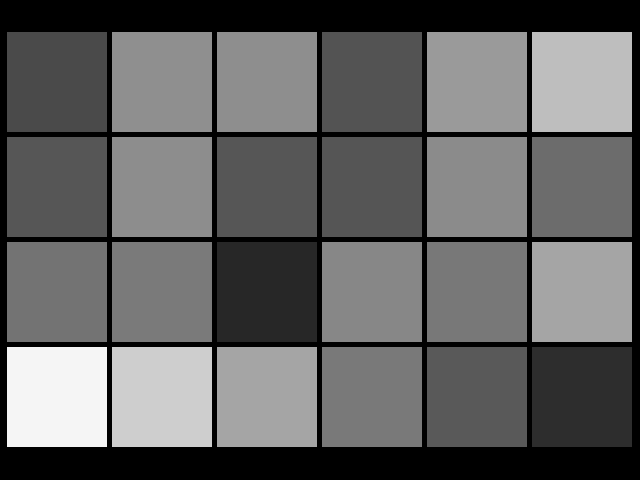}}
  \centerline{(d) Disagreement ($P_{\perp}M$) chart }\medskip
\end{minipage}
\hfill
%
\caption{The input measurements in (a) are white balanced in (b). For row-sum constraint, (c) and (d) show the gamut-limited charts on which the constrained and unconstrained estimators agree and disagree, respectively }
\label{fig:charts}
\end{figure}

For the row-sum constraint, Figure~\ref{fig:charts} uses a chart format to illustrate the raw input measurements, the measurements after white balance, and the projected measurements on which the constrained and unconstrained estimators disagree ($P_{\perp}M$) and agree ($P_{||}M$).  The values of $P_{||}M$ are gamut limited by clipping at $0$. Comparing Fig~\ref{fig:charts}(c) with (a), we see that hue is largely preserved, especially for red and green colors.  Note also that $P_{\perp}M$ consists entirely of shades of gray, on which the constraint is active.  The corresponding plots for the total sum error are exactly the same, since $P_{||}$ and $P_{\perp}$ depend only $F$, which is ${\bf 1}$ for both constraints.

\begin{center}
\begin{table}
\label{tab:fiterr}
    \begin{tabular}{ | l | l | l | l |}
    \hline
    Constraint & Fitting Err & CRLB \\ 
                   & $\|CM-R\|_W$ & $\|J_c^{-1}\|$ \\ \hline 
    None & 0.173 & 20.025 \\ \hline 
    total-sum & 0.191 (+10.4 \%) & 20.009 (-0.08\%) \\ \hline 
    row-sum & 0.202 (+16.8\%) & 19.972 (-0.26\%) \\ \hline 
    \end{tabular}
\caption{Comparison of fitting error and CRLB matrix norms, with row-sum constraint stricter than total-sum. Here $W=I$ and percent changes are calculated with respect to the unconstrained case on the top row.}
\end{table}
\end{center}

\section{SUMMARY AND CONCLUSIONS}

In this paper, we derive exact expressions for the increase in fitting error due to imposing a constraint in linear least squares problems.  We provide descriptions of the subspaces on which the constrained and unconstrained estimators agree.  We illustrate these results with an example from color calibration.

%

\section{APPENDIX}

\subsection{Solution is unique}

We use the constructive approach in \cite{boydvanden18} to prove that the solution (\ref{eq:ourcls}) is the unique minimizer. Let $C$ be such that $C\neq \widehat{C}$ and $KCF=G$.  Then, we write
\begin{flalign}
    \|CM-R\|^2_W &= \|CM-\widehat{C} M + \widehat{C} M - R\|^2_W,\nonumber & &\\
    &= \|CM - \widehat{C} M\|^2_W + \| \widehat{C} M - R\|^2_W \nonumber & & \\
     &+ 2\,{\rm trace}\left\{(CM-\widehat{C} M)W(\widehat{C} M - R)^\top\right\}& &. 
\label{eq:uniquetrace}
\end{flalign}
Writing the rightmost term (inside trace) as
\begin{equation}
    (CM-\widehat{C} M)W(\widehat{C} M - R)^\top = (C-\widehat{C})MW(M^{\top} \widehat{C}^{\top} - R^\top)
    \label{eq:trace}
\end{equation}
From (\ref{eq:normal}), we find that
\begin{equation}
    MW\left(M^{\top}\widehat{C}^{\top} - R^{\top}\right)=-\frac{1}{2}F \Lambda^{\top} K.
\end{equation}
Substituting this back into the trace (\ref{eq:uniquetrace}) leads to 
\begin{equation}
{\rm trace}\left\{(C-\widehat{C})F \Lambda^{\top}K\right\} = 
{\rm trace}\left\{K\left(C-\widehat{C}\right)F\Lambda^{\top}\right\}.
\end{equation}
But $K(C-\widehat{C})F=0$, and consequently, (\ref{eq:trace}) vanishes. This leaves, since $M$ is assumed to have full rank, that $CM-\widehat{C}M\neq 0$, and therefore:
\begin{eqnarray}
    \|CM-R\|^2_W &=& \|CM - \widehat{C} M\|^2_W + \| \widehat{C} M - R\|^2_W \nonumber \\
&>& \| \widehat{C} M - R\|^2_W.
    \label{eq:excess}
\end{eqnarray}

\subsection{Excess fitting error due to constraint}

Using (\ref{eq:ourcls}), we calculate the fitting error with the constrained estimator as follows:
\begin{flalign}
    \|\widehat{C}M-R\|^2_W &= \|C_u M - R - K_{\R} \Delta_c F_{\ell} M \|^2_W,& &  \nonumber \\
    &= \| C_u M - R \|^2_W + \| K_{\R} \Delta_c F_{\ell} M \|^2_W ,\nonumber & &\\
    & + 2\,{\rm trace}\left\{\left(C_u M - R\right)W\left(K_{\R} \Delta_c F_{\ell} M  \right)^\top\right\}. & &
\label{eq:mm}
\end{flalign}
The term in the trace may be expanded as follows. Let 
$Q=K_{\R} \Delta_c F_{\ell}$, and use (\ref{eq:uLS}) to obtain (\ref{eq:excesserror}): 
\begin{flalign}
   & \left(C_u M - R\right)W\left(Q M\right)^\top =  C_u M W M^{\top} Q^{\top} - R W M^{\top} Q^{\top},&&\nonumber\\
    &= RWM^{\top}\left(MWM^{\top}\right)^{-1} \left(MWM^{\top}\right) Q^{\top}
     -  RWM{^\top}Q^{\top},&& \nonumber\\
    &= 0.&&
\end{flalign}

\bibliographystyle{IEEEbib}
\bibliography{icassp_refs}

\begin{thebibliography}{10}

\bibitem{sharma03}
Gaurav Sharma,
\newblock {\em Color imaging handbook},
\newblock CRC Press, Boca Raton, 2003.

\bibitem{macbeth}
``{Color Checker chart},'' \url{https://en.wikipedia.org/wiki/ColorChecker},
\newblock [Online; accessed 29-June-2021].

\bibitem{bianco2013}
Simone Bianco, Arcangelo Bruna, Filippo Naccari, and Raimondo Schettini,
\newblock ``Color correction pipeline optimization for digital cameras,''
\newblock {\em Journal of Electronic Imaging}, vol. 22, pp. 3014--, 04 2013.

\bibitem{fang2017}
Fufu Fang, Han Gong, Michal Mackiewicz, and Graham Finlayson,
\newblock ``Colour correction toolbox,''
\newblock in {\em Proceedings of 13th AIC Congress 2017}. Korea Society of
  Color Studies, Jeju, Korea, October 2017.

\bibitem{boydvanden18}
Stephen Boyd and Lieven Vandenberghe,
\newblock {\em Introduction to Applied Linear Algebra},
\newblock Cambridge University Press, Cambridge, UK, 2018.

\bibitem{amemiya85}
Takeshi Amemiya,
\newblock {\em Advanced Econometrics},
\newblock Harvard University Press, Cambridge, MA, 1985.

\bibitem{finlaysondrew97}
Graham Finlayson and Mark~S. Drew,
\newblock ``Constrained least-square regression in color spaces,''
\newblock {\em Journal of Electronic Imaging}, vol. 6, no. 4, pp. 484--493,
  1997.

\bibitem{wolftechreport}
Stephen Wolf,
\newblock ``Color correction matrix for digital still and video imaging
  systems,''
\newblock Tech. {R}ep. TM-04-406, National Telecommunications and Information
  Administration, 2003.

\bibitem{gormanhero}
John~D. Gorman and Alfred~O. Hero,
\newblock ``Lower bounds for parametric estimation with constraints,''
\newblock {\em IEEE Transactions on Information Theory}, vol. 36, no. 6, pp.
  1285--1301, 1990.

\bibitem{bvconvex}
Stephen Boyd and Lieven Vandenberghe,
\newblock {\em Convex Optimization},
\newblock Cambridge University Press, 2004.

\bibitem{Gross03}
Jurgen Gross,
\newblock {\em Linear Regression},
\newblock Springer-Verlag, Berlin, 2003.

\bibitem{pascale}
Danny Pascale,
\newblock ``{RGB} coordinates of the macbeth colorchecker,''
\newblock Tech. {R}ep., The BabelColor Company, 2006.

\end{thebibliography}

\end{document}